\documentclass[10pt, a4paper]{article}

\usepackage{lrec-coling2024} 
\usepackage[utf8]{inputenc} 
\usepackage[T1]{fontenc}    
\usepackage{amsmath}
\usepackage{booktabs}       
\usepackage{float}
\usepackage{makecell}
\usepackage{paralist}
\newcommand{\setfootnotemark}{%
  \refstepcounter{footnote}%
  \footnotemark[\value{footnote}]}

\title{Training BERT Models to Carry Over a Coding System Developed on One Corpus to Another}

\name{Dalma Galambos$^1$, Pál Zsámboki$^2$} 

\address{$^1$Pázmány Péter Catholic University \\
$^2$HUN-REN Alfréd Rényi Institute of Mathematics \\
$^2$Institute of Mathematics, Faculty of Science, Eötvös Loránd University \\
Budapest, Hungary \\
         galambos.dalma@gmail.com, zsamboki.pal@renyi.hu\\
         }

\abstract{
This paper describes how we train BERT models to carry over a coding system developed on the paragraphs of a Hungarian literary journal to another. The aim of the coding system is to track trends in the perception of literary translation around the political transformation in 1989 in Hungary. To evaluate not only task performance but also the consistence of the annotation, moreover, to get better predictions from an ensemble, we use 10-fold crossvalidation. Extensive hyperparameter tuning is used to obtain the best possible results and fair comparisons. To handle label imbalance, we use loss functions and metrics robust to it. Evaluation of the effect of domain shift is carried
out by sampling a test set from the target domain. We establish the sample size by estimating the bootstrapped confidence interval via simulations. This way, we show that our models can carry over one annotation system to the target domain. Comparisons are drawn to provide insights such as learning multilabel correlations and confidence penalty improve resistance to domain shift, and domain adaptation on OCR-ed text on another domain improves performance almost to the same extent as that on the corpus under study. See our code at \url{https://codeberg.org/zsamboki/bert-annotator-ensemble}.
 \\ \newline  }

\begin{document}

\maketitleabstract

\section{Introduction}

\subsection{Objective of the Large Pilot Project Providing the Broader Context of the Present Paper} \label{ss:broader project}

From the aspect of cultural policy, transition from the Socialist Kádár era (1956–1989) to democracy in Hungary was a crucial period in time. Culture, particularly literature and by extension, literary translation had been heavily funded by the state before the so-called political transformation in 1989, until which literary translators, consequently, had enjoyed a much higher status than in the period since.

This large pilot project chooses a data-driven path to examine this change and blends qualitative and quantitative methods in order to provide a closer look at how literary translators were perceived in the two decades surrounding the regime change. It utilizes a new coding system (we also refer to this as \emph{annotation system} in our paper) tailored to the domain, state-of-the-art classification technology, quantitative and qualitative analysis and network analysis. Background of the project in literary translation studies as well as a more detailed account of the manual coding process and results are discussed in the doctoral dissertation of \citet{galambos2023muforditas}.

\subsection{Scope of the Present Paper, Main Contributions}

The present paper details the classification technology that we use. Since their discovery, transformers \citep{vaswani2017attention} have been dominating the Natural Language Processing (NLP) field. For classification, the BERT architectures \citep{devlin2019bert} are widely and successfully used. We train BERT models on a manually annotated dataset to apply the coding system to another domain, which we call the \emph{target domain}.

Our main contributions discussed in this paper are as follows:
\begin{enumerate}
    \item We show that with extensive hyperparameter tuning both in pretraining (\S\ref{subsection: domain adaptation}) and finetuning, and with loss functions robust to label imbalance in the latter (\S\ref{subsection: finetuning}), we can teach BERT models complex and highly imbalanced sequence labelling systems. This is verified via 10-fold crossvalidation, the resulting models forming model ensembles for prediction.
    \item To evaluate the resistance of our models to domain shift, we select a test set from the target domain for manual validation. We introduce a method to estimate confidence intervals of test results with various sample sizes. We verify that our models can carry over one coding system to the target domain (\S\ref{subsection:validation on the predict set}).
    \item In addition to finetuning off-the-shelf Hungarian BERT models, and ones pretrained on the corpus under study, we also finetune models pretrained on OCR-ed text of similar layout and typography from the same time period but with very different subject matter. We show that adaptation to the peculiarities of the OCR-ed text without the domain knowledge gives almost as much improvement on an off-the-shelf model as adaptation to the corpus under study (\S\ref{subsubsection:domain adaptation comparison}).
    \item We run further comparisons with different loss functions, and numerous low-cost baseline methods (\S\ref{section:comparisons}). First of all, we show that transformers have a clear advantage over low-cost baselines based on bag-of-words and word embedding. We point out further tendencies such as a multilabel classifier is more resistant to domain shift than individual binary classifiers, and adding confidence penalty to the BERT finetuning loss also has a beneficial effect in domain change.
\end{enumerate}

\subsection{Related Work}

Training word embeddings on a corpus from the journal \emph{Pártélet}, the official journal of the governing party in Hungary in the Kádár era, \citet{ring2020kulcsfogalmak} study trends in the semantic changes of notions related to decisions and control, while \citet{szabo2021exploring} perform a similar study for notions related to agriculture and industry.

BERT has been successfully used to learn and predict complex sequence labelling systems in several domains. \citet{bressem2020highly} train models on an annotated set of chest radiology reports. They show that their best model can then predict labels on CT reports. \citet{grandeit2020using} train models on counselling reports. They conclude that out of A) the labels predicted by their best model, B) the labels given by an expert annotator and C) the labels given by a novice annotator, A) and B) are the most similar pair. \citet{limsopatham2021effectively} trains models on legal documents. Out of the solutions he tested, Longformer \citep{beltagy2020longformer} is shown to give the best results when being taught on long sequences. \citet{mehta2022psychotherapy} train models on therapist talk-turns. They show that even when their best model cannot always correctly classify the approach used in each talk-turn, it can still reliably tell which approaches have been used during a therapy session.

With regards to measuring the impact of OCR-ed text on NLP task performance: \citet{jiang2021evaluating} and \citet{jiang2021impact} compare BERT contextual token embeddings on pairs of the cleaned text (Guthenberg) and OCR-ed text (HathiTrust) of the same books. Both studies find that pretraining either on clean or OCR-ed text helps performance. \citet{labusch2020named} perform Named Entity Recognition and Linking on OCR-ed documents kept at the Berlin State Library. They find that pretraining BERT on historical text worsens task performance on contemporary text.

\section{Dataset}

\subsection{Corpus: \emph{Alföld} and \emph{Nagyvilág}, Two Hungarian Literary Journals from the Period under Examination}\label{subsection: corpus}

When it comes to examining the status of literary translators and translation, \emph{Nagyvilág} is the single most significant journal of the Kádár era its primary focus being on world literature and related articles. On the other hand, its scope makes any in-depth longitudinal analysis a resource-intensive task to carry out. Which is why the training set was retrieved from the journal \emph{Alföld}, that is somewhat less relevant to literary translation, as it predominantly features Hungarian literature and related articles. The page scans of these journals, as well as those used in the domain adaptation comparisons (\S\ref{subsubsection:domain adaptation comparison}) were downloaded from the Arcanum database \citeplanguageresource{arcanum}.

\subsection{Manual Annotation of \emph{Alföld}}\label{subsection: manual annotation}
The training set consists of a manually annotated database listing all paragraphs from \emph{Alföld} (with the exception of pieces of or excerpts from literary works) that mention translation to any extent thematically annotated with two kinds of labels.
\begin{enumerate}
    \item Content labels indicate what implications, connotations, themes or topics are touched upon in each paragraph in reference to translation. Each paragraph may be coded with several content labels (multilabel coding). 38 content labels are used.
    \item Context labels however signal what it is in the context that warrants mentioning translation (e.g.~the paragraph is about the author of a book that was translated, etc.). Each paragraph may be coded with only one context label (multiclass coding). 11 context labels are used.
\end{enumerate}
It is important to clarify here that even though the two labeling systems hold some similarly named labels, the two systems are drastically different in principle. Content labels show \emph{what themes are mentioned} in a paragraph relating to translation and context labels show \emph{why translation is mentioned} in a paragraph in the first place. As examples and because we use them in the qualitative analysis (\S\ref{ss:qualitative analysis}), we give the definitions of the content label \emph{author as translator} and the context label \emph{translator} in Subsection \ref{ss:label definitions}.

The project being in its pilot phase, the system and list of codes are developed and annotation is conducted by Galambos to create the training set during the first phase of the project. Content analysis and certain features of thematic analysis \citep{braun2022thematic} are combined to achieve as accurate and unbiased results as possible considering that all coding system adjustments and annotation are implemented by a single researcher. For this purpose, annotation is performed twice with a significant time gap between the two iterations. This helps finetuning the coding system and eliminating inconsistencies and other mistakes. Deploying only one annotator at this phase is also one of the reasons for seeking rigorous validation options, as seen in Sub-subsection \ref{subsubsection:10-fold training} and Section \ref{subsection:validation on the predict set}.

Despite its obvious advantages regarding robustness, at this stage, using several annotators was not an option, mainly due to the project’s experimental nature and given that creating the coding system was an ongoing part of the annotation process itself to test certain hypotheses about eliminating biases by not establishing a fixed set of labels beforehand but rather making the identification of labels part of the annotation phase. A potential next stage would involve further changes to the method and again, working from the bottom up with improved conditions and, undoubtedly, more annotators, heavily building on the conclusions of the pilot stage to eliminate disadvantages we have identified.

\subsection{Preprocessing Pipeline: from Page Scans to Paragraph Texts}\label{subsection: preprocessing}

We need to transform the \emph{Alföld} and \emph{Nagyvilág} journal scans and the annotated paragraphs from \emph{Alföld} to a form that a Large Language Model (LLM) can process. To this end, the scanned journal pages first need to be transformed to a sequence of paragraph texts.

We use the Tesseract \citep{tesseract} OCR engine via the Python Tesseract \cite{pytesseract} interface. It can accurately split pages to paragraphs. However, we also need to recognize cases when a page break is also a paragraph break.

For that purpose, we apply the DBSCAN clustering algorithm \citep{ester1996density} via its Scikit-learn \citep{scikit-learn} interface to bounding box statistics. We can use this to determine 
\begin{enumerate}
    \item the type of a paragraph such as main text, footnote, and heading, and
    \item whether the horizontal coordinates of a line suggest that it is the first or last line of a paragraph.
\end{enumerate}

We then match the paragraphs resulting from this pipeline with the quotes in the annotation dataset using a bag-of-words-based distance. It is verified by hand that the only matching errors come from occasionally incorrectly separating paragraphs.

\subsection{Dataset Statistics and Further Transformations}\label{subsection: dataset statistics}

\subsubsection{Paragraph and Word Counts}\label{subsubsection:paragraph and word counts}

Via the preprocessing pipeline described in Subsection \ref{subsection: preprocessing}, we collect 9,619,240 words in 206,921 paragraphs from the \emph{Alföld} issues of 1980--1999, and 11,622,881 words in 322,970 paragraphs from the \emph{Nagyvilág} issues of 1980--1999. Therefore, for domain adaptation we can use a dataset with 21,242,121 words.

\subsubsection{Pruning \emph{Alföld} for the Finetuning Set}

Out of the 206,921 paragraphs in the \emph{Alföld} issues from the years 1980--1999, only 1515, that is 0.73\% concern translation. On the other hand, out of these 1515 paragraphs, 1467, that is 96.83\% contain the subword ``fordí'' (a fragment of the word ``translation'' in Hungarian which is ``fordítás''). Therefore, by restricting the train set to the 3994 paragraphs that contain the subword ``fordí'', we can discard a vast amount of unneeded data while losing only a handful of relevant entries.

A further restriction comes from the architecture: in January 2023, when setting up training, there is no Hungarian LLM to our knowledge that was pretrained on suitably long sequences. Based on our preliminary experiments, we choose PULI-BERT-Large \citep{yang2023jonnek}, which we use via the Huggingface Transformer library \cite{wolf2020transformers}. This model has a maximum token size of 512. Therefore, we restrict the train set to sequences of 512 tokens at most. This results in a finetuning train set of 3134 sequences. Out of these 3134 paragraphs, 1975 do not concern translation. The main reason should be the fact that the Hungarian word for translation also has other unrelated meanings. To handle these cases, we use an ``unrelated'' context label.

\subsubsection{Label Statistics}

\begin{figure*}[!ht]
    \centering
    \includegraphics[scale=0.32]{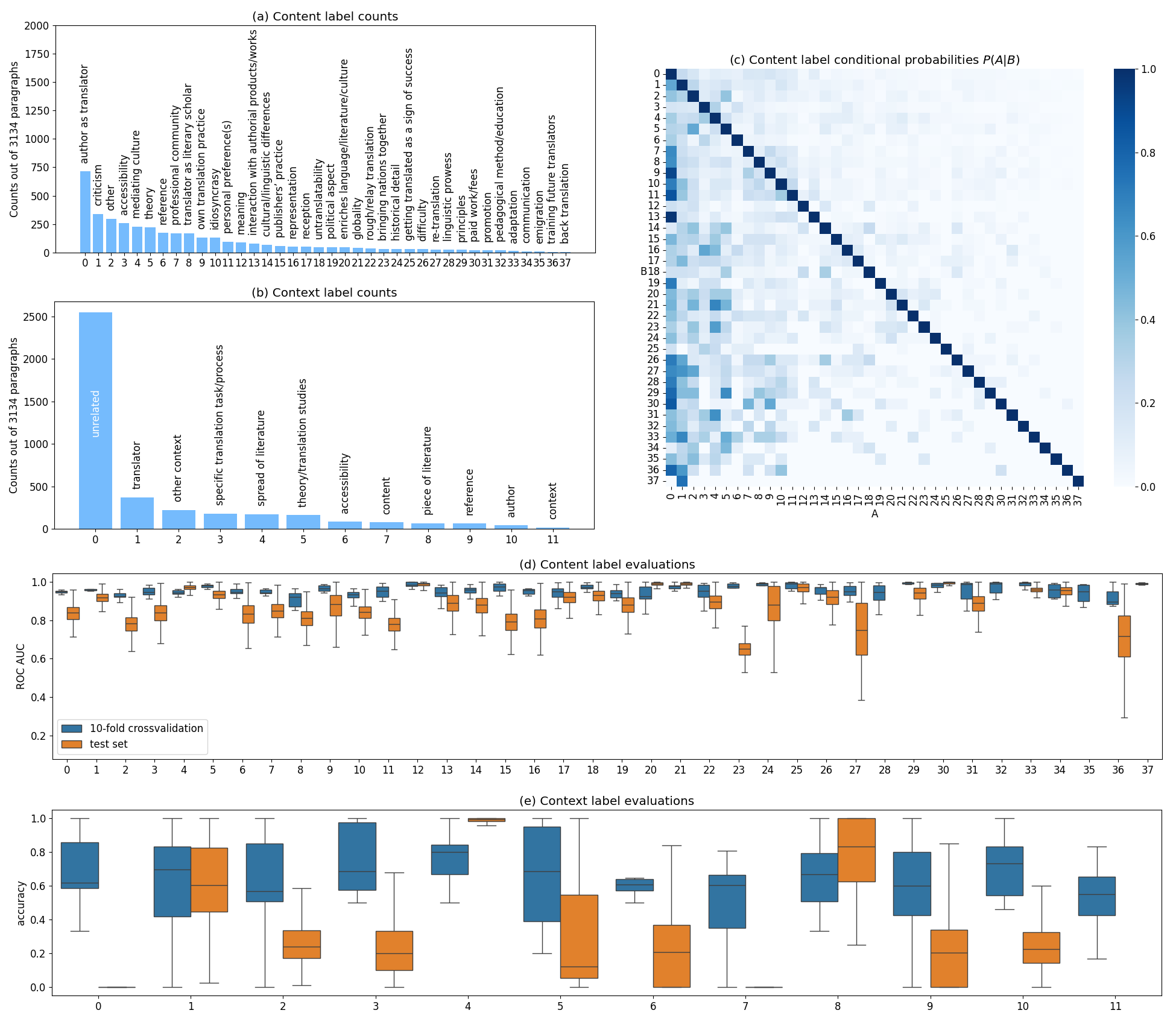}
    \caption{Dataset statistics and evaluation results. (a) Content label counts. (b) Context label counts. The context label with index 0 refers to paragraphs that contain the subword ``fordí'' but are unrelated to translation. (c) Content label correlations expressed as conditional probabilities. (d) Content label evaluation results by label (ROC AUC). (e) Context label evaluation results by label (accuracy). In (d) and (e), 10-fold crossvalidation results are dark blue, and test set results are orange.}
    \label{fig:diagrams}
\end{figure*}

Both the content and context labels are highly imbalanced. The mean imbalance ratio \citep[\S3.1]{charte2013first}, that is the average ratio of the maximal label count to label counts, is $27.34$ for content labels and $36.31$ for context labels. For content labels, there is further imbalance in the imbalance ratios of concurrent labels: the SCUMBLE score \citep[\S3.2]{charte2014concurrence}, that is the mean Atkinson index \citep{atkinson1970measurement} of imbalance ratios of labels present in a paragraph, is $0.3290$. For more details, see Figure \ref{fig:diagrams}a-c. 

\subsubsection{10-fold Stratification}

We use stratification to get both the content and context label 10-folds. This is straightforward in the case of context labels, but not so for content labels. Following \citep{sechidis2011stratification}, we seek to find a partition where the individual label frequencies approximate those on the full dataset. Making use of GPU parallelization, we draw millions of partitions and choose one that is minimal in the reverse lexicographical ordering of individual label frequency error rates normalized by individual label frequencies.

\subsubsection{Pruning and Truncating Paragraphs from \emph{Nagyvilág} for the Target Domain}\label{subsubsection:pruning Nagyvilag}

As for forming the target domain from the preprocessed \emph{Nagyvilág} corpus, out of its 322,970 paragraphs, we select the 12,712 that contain the subword "fordí". The dataset is further filtered by discarding (i) tables of contents, (ii) references at the end of quotations, literary texts or other articles that only consist of ``translated by <translator>'' and (iii) literary texts. This way, we end up with a target domain of 4589 sequences. We truncate the tokenized sequences at 512 tokens in front. See Section \ref{subsection:validation on the predict set} to see how we use importance sampling to get a test set from this, which we then use to check the resistance of our models to domain shift.

\section{Training}\label{section:training}

\subsection{Pretraining: Domain Adaptation}\label{subsection: domain adaptation}

We perform domain adaptation with Masked Language Modelling on the 21,242,121-word \emph{Alföld}--\emph{Nagyvilág} dataset described in Sub-subsection \ref{subsubsection:paragraph and word counts}. The batch size is set to the largest value that fit in the NVIDIA A100 40GB GPU that we are using, and following \citet{ma2021adequacy}, the number of warmup steps is set to $\frac{2}{1-\beta_2}$ train steps, where $\beta_2$ denotes the second momentum in the AdamW optimizer \citep{loshchilov2019decoupled}. See Table \ref{table:hyperparameter initial distributions} in the Appendix for the hyperparameters tuned. Training with the best hyperparameters that has been found brings down the perplexity score of 43.07 of the original PULI-BERT-Large model to 2.88. The cause of the magnitude of this decrease could be that the 21 million-word domain adaptation corpus is rather small in comparison to the 86 billion-word corpus the model was originally pretrained on \citep[Table 1 (``1. t\'abl\'azat'')]{yang2023jonnek}.

\subsection{Finetuning: Imbalanced Label Classification}\label{subsection: finetuning}

As described in Subsection \ref{subsection: dataset statistics}, we work with a 3134-sequence finetuning train set with two highly imbalanced label sets: 38 content labels that are multilabel, and 12 context labels that are single label.

\subsubsection{10-fold Training and Evaluation}\label{subsubsection:10-fold training}

The small size of the train set is taken advantage of by using techniques requiring several train runs. One of these is 10-fold training. It offers 3 main benefits:
\begin{enumerate}
    \item Consistency of evaluation scores across iterations confirm consistency of annotation.
    \item More robust evaluation scores can be achieved with confidence intervals.
    \item The 10 models acquired from training can be used for inference as an ensemble.
\end{enumerate}

\subsubsection{Population-Based Training}\label{subsubsection:population-based training}

The other technique with several runs we use is the application of Population-Based Training \citep{jaderberg2017population} for hyperparameter optimization. Its benefits are 2-fold:
\begin{enumerate}
    \item It adapts hyperparameters on the fly and this way finds hyperparameter schedules on its own.
    \item Since it trains the samples in parallel, it is highly scalable.
\end{enumerate}
We perform this algorithm by training 100 models in parallel, epoch by epoch. We train them for 30 epochs for the content labels, and at least 30 epochs until there is no improvement for 10 epochs for the context labels. In our version of selection:
\begin{enumerate}
    \item the top 10\% elite is kept unchanged, and
    \item roulette wheel selection is used for the rest with hyperparameter perturbation. 
\end{enumerate}
In contrast to \citet{jaderberg2017population}, in perturbation, we do not choose between the multipliers 0.8 and 1.2, but pick a multiplier uniformly from the interval $[0.8, 1.2]$. We make these changes to have less rigid heuristics.

See Table \ref{table:hyperparameter initial distributions} in the Appendix for the hyperparameters tuned.

\subsubsection{Content Label Finetuning}\label{subsubsection:content label finetuning}

In order to avoid fixing a threshold, and for its robustness to label imbalance, we use macro averaged ROC AUC as evaluation metric. Moreover, we use focal loss \citep{lin2017focal} as train loss. This results in a very satisfactory average ROC AUC of $0.9524\pm0.0114$ (we report all of the confidence intervals with confidence level 95\%). See more detailed evaluation results in Figure \ref{fig:diagrams}d. Note that the less frequent labels do not get lower scores. We use the unweighted version \citep[Equation 4]{lin2017focal}, as the weighted one \citep[Equation 5]{lin2017focal} does not bring any improvement.

\subsubsection{Context Label Finetuning}\label{subsubsection:context label finetuning}

The focus being resistance to label imbalance here as well, we use balanced accuracy as evaluation metric. As training loss, we combine label distribution-aware margin loss \citep{cao2019learning} with a penalty for confident output distributions \citep{pereyra2017regularizing}. This gives a balanced accuracy of $0.6357\pm0.1266$. See Figure \ref{fig:diagrams}e for more detailed evaluation results, and Sub-subsection \ref{subsubsection:context label losses} for comparisons with other loss functions.

Note that separate model ensembles are trained in the content and the context label case. We experiment with training combined models for the two label sets, but with significantly worse results.

\section{Evaluation on the Target Domain}\label{subsection:validation on the predict set}

To measure if our model ensemble can successfully carry over the two coding systems to the target domain, we draw a sample from it. Manually annotating this, we get a test set. Note that our models never see the labels on the test set.

\subsection{Deciding the Size of the Test Set}

\begin{figure}[!ht]
    \centering
    \includegraphics[scale=0.5]{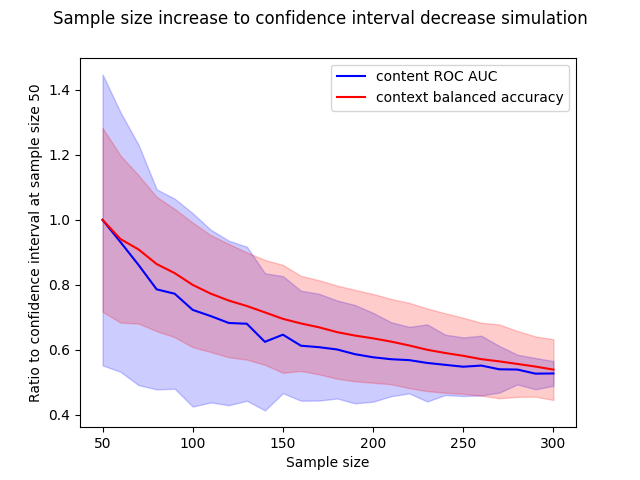}
    \caption{Sample size increase to confidence interval decrease simulation via bootstrap.}
    \label{fig:sample size increase to confidence interval decrease}
\end{figure}

As manual validation is highly resource-intensive, when deciding on how many samples we should draw for the test set from the target domain, for a prospective sample size, we seek to estimate what confidence interval is to be expected. To that end, simulations are run on the results of the 10-fold crossvalidation: for both content and context labels, $\mathtt{sample\_size}=50, 60, \ldots, 300$ and each fold evaluation set, 100 times, we draw a random sample of \texttt{sample\_size} from the fold evaluation set, and via bootstrapping with size 10,000 the standard deviation of the relevant metric is approximated, see Figure \ref{fig:sample size increase to confidence interval decrease}. This we can use to estimate what confidence interval we would get from what sample size. Based on this data, we determine that going from 50 samples to 100 for a confidence interval decrease of about 20\% is worthwhile, however, going up to 150 for a further confidence interval decrease of about 10\% is not. Therefore a decision is made to draw 100 samples from the target domain for manual validation.

\subsection{Importance Sampling on the Target Domain}

As we naturally do not have labels for the target domain, only model prediction probabilities, a stratified sample for the test set cannot be used. Therefore, to address label imbalance, we opt for an importance sampling approach, that draws paragraphs with less frequent label prediction probabilities with higher probability.

\subsection{Validation Results}

On the content label set, we achieve a ROC AUC of $0.8757 \pm 0.0252$ (to get confidence intervals and box plots for the validation results, we use bootstrapping on 10,000 samples). For more detailed scores, see Figure \ref{fig:diagrams}d. This means that the model ensemble is capable of reliably carrying over the labeling to the target domain.

On the other hand, on the context label set, we only reach a balanced accuracy of $0.3287 \pm 0.1297$. For more detailed scores, see Figure \ref{fig:diagrams}e. We perform a preliminary investigation on the possible reasons for misclassification in Subsection \ref{ss:qualitative analysis}. Based on this result, from \emph{Nagyvilág}, only content labels are used in the findings.

\section{Comparisons}\label{section:comparisons}

\begin{table*}[!p]
    \centering
    \begin{tabular}{cccccc}
        \toprule
        & & \multicolumn{2}{c}{Content labels (ROC AUC)} & \multicolumn{2}{c}{Context labels (balanced accuracy)} \\
        \cmidrule(r){3-6}
        Classifier & \makecell{Re-\\sampler} & \makecell{10-fold \\ crossvalidation} & \makecell{test set}
        & \makecell{10-fold \\ crossvalidation} & \makecell{test set} \\
        \midrule
        \multicolumn{6}{c}{Based on TF-IDF vectors} \\
        \midrule
        CatBoost\setfootnotemark\label{fourth} & ADASYN & & & $0.3372 \pm 0.0719$ & $0.1365 \pm 0.1027$ \\
        CatBoost & MLSOL & $0.8144 \pm 0.0535$ & $0.6822 \pm 0.0376$ \\
        CatBoost & R-HwR\setfootnotemark\label{first} & $0.8100 \pm 0.0496$ & $\mathbf{0.6937 \pm 0.0375}$ \\
        CatBoost & SMOTE & & & $0.3469 \pm 0.0951$ & $0.1188 \pm 0.1015$ \\
        LightGBM & ADASYN & & & $0.3433 \pm 0.0873$ & $0.1180 \pm 0.0876$ \\
        LightGBM & SMOTE & & & $0.3454 \pm 0.0720$ & $0.1737 \pm 0.1068$ \\
        XGBoost\setfootnotemark\label{third} & ADASYN & $\mathbf{0.8530 \pm 0.0255}$ & $0.6703 \pm 0.0347$ & $\mathbf{0.3629 \pm 0.0761}$ & $\mathbf{0.1939 \pm 0.1156}$ \\
        XGBoost & SMOTE & & & $0.3500 \pm 0.1025$ & $0.1432 \pm 0.0951$ \\
        \midrule
        \multicolumn{6}{c}{Based on fastText vectors} \\
        \midrule
        CatBoost & ADASYN & & & $0.3309 \pm 0.0495$ & $0.1573 \pm 0.1014$ \\
        CatBoost & MLSOL & $0.8475 \pm 0.0418$ & $0.7594 \pm 0.0313$ \\
        CatBoost & R-HwR & $0.8412 \pm 0.0396$ & $\mathbf{0.7797 \pm 0.0318}$ \\
        CatBoost & SMOTE & & & $\mathbf{0.3345 \pm 0.0605}$ & $0.1430 \pm 0.1029$ \\
        fastText & & $0.7330 \pm 0.0383$ & $0.6753 \pm 0.0439$ & $0.2536 \pm 0.0607$ & $\mathbf{0.1780 \pm 0.1204}$ \\
        LightGBM & ADASYN & & & $0.3157 \pm 0.0462$ & $0.1308 \pm 0.0821$ \\
        LightGBM & SMOTE & & & $0.3274 \pm 0.0855$ & $0.1530 \pm 0.0937$ \\
        XGBoost & ADASYN & $\mathbf{0.8702 \pm 0.0256}$ & $0.7176 \pm 0.0363$ & $0.3333 \pm 0.0983$ & $0.1336 \pm 0.0677$ \\
        XGBoost & SMOTE & & & $0.3162 \pm 0.0816$ & $0.1330 \pm 0.0971$ \\
        \midrule
        \multicolumn{6}{c}{BERT training methods} \\
        \midrule
        Model & \makecell{Domain\\Adaptation} & \multicolumn{2}{c}{\makecell{Content label results \\ with Focal Loss}} & \multicolumn{2}{c}{\makecell{Context label results \\ with LDAM + CP loss}}\\
        \midrule
        huBERT & Corpus & $0.9503 \pm 0.0175$ & $0.8727 \pm 0.0268$ & $0.6173 \pm 0.0872$ & $0.3032 \pm 0.1422$ \\
        huBERT & Extended & $0.9468 \pm 0.0064$ & $0.8702 \pm 0.0362$ & $0.6187 \pm 0.0486$ & $0.3255 \pm 0.1285$ \\
        huBERT & OCR & $0.9478 \pm 0.0113$ & $0.8737 \pm 0.0268$ & $0.6004 \pm 0.1147$ & $0.2788 \pm 0.1175$ \\
        huBERT & None & $0.9093 \pm 0.2135$ & $0.8711 \pm 0.0263$ & $0.5584 \pm 0.1227$ & $\mathbf{0.3671 \pm 0.1355}$ \\
        PULI\setfootnotemark\label{second} & Corpus & $0.9524 \pm 0.0114$ & $0.8757 \pm 0.0252$ & $0.6357 \pm 0.1266$ & $0.3287 \pm 0.1297$ \\
        PULI & Extended & $\mathbf{0.9534 \pm 0.0065}$ & $\mathbf{0.8808 \pm 0.0312}$ & $\mathbf{0.6410 \pm 0.0481}$ & $0.2672 \pm 0.1332$ \\
        PULI & OCR & $0.9481 \pm 0.0139$ & $0.8767 \pm 0.0252$ & $0.6265 \pm 0.0842$ & $0.3560 \pm 0.1487$ \\
        PULI & None & $0.9193 \pm 0.0673$ & $0.8513 \pm 0.0280$ & $0.5545 \pm 0.1422$ & $0.2752 \pm 0.1416$ \\
        \midrule
        \makecell{Domain \\ Adapted \\ Model} & Loss \\
        \midrule
        PULI & Focal & & & $0.5985 \pm 0.0991$ & $0.3428 \pm 0.1431$ \\
        PULI & LDAM & & & $0.6263 \pm 0.0938$ & $0.2382 \pm 0.1356$ \\
        \midrule
        \multicolumn{3}{l}{Double Context Length Domain Adapted Model} \\
        \midrule
        huBERT & & & & $0.6048 \pm 0.0318$ & $0.3460 \pm 0.1312$ \\
        \midrule
        \multicolumn{3}{l}{Single Model with no Domain Adaptation} \\
        \midrule
        huBERT & & & & $0.6029$ & $0.3637 \pm 0.1365$ \\
        PULI & & & & $0.5751$ & $0.3471 \pm 0.1307$ \\
        Llama 2 & & & & $0.4009$ & $0.2605 \pm 0.1296$ \\
        \bottomrule
    \end{tabular}
    \caption{Comparisons on baseline models and BERT training methods. $^1$In the case of CatBoost on content labels, a single multilabel classifier is tuned. $^2$Short for REMEDIAL-HwR. $^3$In the case of XGBoost on content labels, individual binary classifiers are tuned for each label. $^4$Short for PULI-BERT-Large.}
    \label{fig:comparisons}
\end{table*}

\subsection{Baseline Methods}

We test baseline methods with low computational cost. We discuss transforming paragraphs to tabular data in Sub-subsection \ref{sss:baseline paragraphs to tabular}, resampling algorithms in Sub-subsection \ref{sss:baseline resampling}, gradient boosted tree algorithms in Sub-subsection \ref{sss:baseline gbt} and  All of the options are extensively tuned using the BlendSearch algorithm.

\subsubsection{Transforming Paragraphs to Tabular Data}\label{sss:baseline paragraphs to tabular}

As most low-cost machine learning methods work on tabular data, we first need to transform the paragraphs to numerical vectors. For a bag-of-words-based approach, we test using TF-IDF vectors, via the \texttt{scikit-learn} \citep{scikit-learn} implementation. For a low-cost word embedding-based approach, we test using the sentence vectors output by the fastText word representation model \citep{bojanowski2017enriching}, pretrained on Hungarian Common Crawl and Wikipedia, that is available on their webpage \citep{fasttextpretrained}. In both cases, dimension reduction is performed using the TruncatedSVD \citep{halko2011finding} algorithm. We also test the built-in fastText classifier \citep{joulin2017bag}.

\subsubsection{Resampling the Train Set to Make it more Balanced}\label{sss:baseline resampling}

To make the train set more balanced, we use resampling. On the context label dataset, we test the ADASYN \citep{haibo2008adaptive} and SMOTE \citep{chawla2002synthetic} synthetic oversampling algorithms, via their \texttt{imbalanced-learn} \citep{lemaitre2017imbalanced} implementation. On the content label dataset, we test the REMEDIAL-HwR \cite{charte2019tackling} and MLSOL \cite{liu2020synthetic} synthetic resampling algorithms. We change both algorithms:

REMEDIAL-HwR is in fact the composition of the REMEDIAL \citep[Algorithm 4]{charte2019tackling} and MLSMOTE \citep[Algorithm 1]{charte2015mlsmote} algorithms. In REMEDIAL, entries with a SCUMBLE score higher than a tuneable threshold are selected for decoupling. In MLSOL on the other hand, entries with an IRLbl score larger than the mean are selected to serve as synthetic instance sources. We replace the mean IRLbl score with a tuneable threshold.

The inbalance metric of MLSOL is based on the \emph{local imbalance matrix} $C_{ij}$ \citep[Equation 1]{liu2020synthetic}. To decide the labels of the synthetic entries, a threshold $\theta$ is used (\emph{ibid}., Algorithm 3, line 17). In the paper, this threshold is determined by hardcoded rules (\emph{ibid}., lines 12-16). We let the threshold linearly depend on local label imbalance: $\theta=\frac{1 + C_{ij}}{2}$.

\subsubsection{Using Gradient Boosted Tree Algorithms on the Resampled Train Set}\label{sss:baseline gbt}

Based on their performance in the low-cost domain, we train CatBoost \citep{prokhorenkova2018catboost}, LightGBM \citep{ke2017lightgbm} and XGBoost \citep{chen2016xgboost} models on the resampled context, that is multiclass dataset. As by 2023 October only CatBoost has stable multilabel classification support, we only use that to train multilabel models. We moreover pick the combination that performs best on context labels: XGBoost + ADASYN, to train 38 individual binary classification models on each content label. Note that in the latter case tuning is also performed separately for each binary classifier.

\subsubsection{Discussion of Baseline Results}\label{sss:baseline results}

On content labels, we see an advantage of word embedding (fastText) vectors over bag-of-words (TF-IDF) ones. This could be attributed to the fact that content labels are based on more local information. On the other hand, on context labels, we have a somewhat better result using bag-of-words. This could be due to the tendency that the information expressed in a bag-of-words vector, albeit more reduced, is more balanced in terms of influence by individual words.

As content labels are multilabel, one can also compare training individual binary classifiers, one for each label, to training only one multilabel classifier. We use XGBoost and CatBoost for the two respective approaches. Whichever of bag-of-words or word embedding vector-based feature vectors we use, on source domain 10-fold crossvalidation, one can notice a slight advantage of training individual binary classifiers, but on the target domain test set, one can observe a more pronounced advantage of training a unique multilabel classifier. Learning label correlations may help robustness.

The fastText classifier has significantly worse results in all respects besides context label results on the target domain test set.
\subsection{BERT Training Methods}

In this subsection, we detail different aspects of the training that we test with a number of options. As the best training procedure has already been described in Section \ref{section:training}, here we discuss results right after explaining the component that we change, and not in a separate sub-subsection. 

\subsubsection{Pretrained Models}\label{subsubsection:pretrained models}

We test two Hungarian pretrained models. The earlier model, huBERT \citep{nemeskey2021introducing} has a BERT-Base architecture, and it was trained for 189,000 steps on the Hungarian WebCorpus 2.0 \citeplanguageresource{nemeskey2020natural}, that was built from Common Crawl and includes a little over 9 billion words. PULI-BERT-Large \citep{yang2023jonnek} has a BERT-Large architecture, and it was trained for 750,000 steps on a corpus assembled from the Hungarian WebCorpus 2.0, the Hungarian Wikipedia and a number of other resources, totalling 36,285,941,699 words. The two models seem to perform very similarly on content labels. This seems to indicate that for the content label classification problem a BERT-Base model is big enough. On context label 10-fold results on the other hand, we see a slight advantage of the larger model. The context label test set results appear noisy, without further study, a clear explanation does not seem possible.

\subsubsection{Domain Adaptation}\label{subsubsection:domain adaptation comparison}

We evaluate finetuning after four different options for domain adaptation:

\begin{asparaenum}
    \item No domain adaptation.
    \item Domain adaptation on the corpus under study, that is the 1980--1999 issues of \emph{Alföld} and \emph{Nagyvilág}.
    \item A corpus of similar size of OCR-ed journals of similar layout and typography from the same time period, but with entirely different subjects. See Section \ref{s:ocr corpus} for the list of the journals.
    \item Domain adaptation on an extended contextual corpus consisting of the 1960--2021 issues of \emph{Alföld} and the 1960--2015 issues of \emph{Nagyvilág}.
\end{asparaenum}

Based on our results, adaptation to domain text gives the most perfomance boost, in particular, more contextual data is yet better. However, adaptation to the peculiarities of OCR-ed text is almost as effective, and still significantly better than no adaptation.

\subsubsection{Context Label Losses}\label{subsubsection:context label losses}

As discussed above, we finetune the domain adapted PULI-BERT-Large on the content label set with Focal Loss \citep{lin2017focal}, and this gives a very satisfactory result. On the other hand, on the context label set we want to see if we can improve our result. Therefore, we test Label Distribution-Aware Margin (LDAM) loss \cite{cao2019learning} with and without a Confident output distribution Penalty (CP) \citep{pereyra2017regularizing}. Based on our results, LDAM gives better 10-fold crossvalidation results already on its own, but CP improves the test set results significantly. This may indicate that the regularization effect of CP helps robustness.

\subsubsection{Double Context Length}

As one of the possible reasons for the inferior performance on context labels is that that task requires a larger context length, we experiment with domain adaptation and finetuning with a modified huBERT model, where instead of the original context length of 512, we use 1024. As (for an undisclosed reason) BERT models use learned positional embeddings, following \citep[\S5]{beltagy2020longformer} to extend these to positions $512,\dotsc,1023$, we copy the 512 embeddings twice. In the end, the results do not improve.

\subsection{Llama 2}

Preliminary studies on the performance of the open family of foundation and chat models Llama 2 \citep{touvron2023llama} are conducted.

\subsubsection{Chain-of-Thought Few Shot Learning}

As even evaluating examples is resource-intensive, we only experiment with the content label \emph{author as translator}. We prompt the model with the task description and some randomly chosen examples with chain-of-thought (CoT) explanations \citep{wei2022chain}, and tell it to do the same for an additional paragraph.

It does generate its answers according to the CoT pattern it received, but the linguistic and factual knowledge required to answer questions of this complexity seem to be missing. We do not observe a difference in this respect between the chat models of different sizes. Again, this is an exploratory experiment. It is quite possible that for example with instruction finetuning, better results can be achieved by a generative model.

\subsubsection{Finetuning on the Context Label Set}

We also try finetuning the smallest, 7-billion-parameter foundation model on the context label set. This is computationally very intensive: Even with QLoRA \citep{dettmers2023qlora} and Distributed Data Parallel in its \texttt{accelerate} implementation \cite{gugger2022accelerate} with 5 NVIDIA A100 40GB GPU an epoch takes 20 times as long as training PULI on 1 GPU.

Therefore, instead of 10-fold crossvalidation with Population-Based Training, we opt for BlendSearch on a single train-test split. See Table \ref{table:hyperparameter initial distributions} for the hyperparameter initial distributions. We run the tuning algorithm for 3 days on the 5 GPUs. For a fairer comparison, we finetune the original huBERT and PULI models for 12 hours on 1 GPU the same way. Still, the BERT models surpass Llama 2.

\subsection{Qualitative Analysis}\label{ss:qualitative analysis}

We go through the test set and try to explain from the text the misclassifications of the content label \emph{author as translator} and the context label \emph{translator} by the PULI model domain adapted on the main corpus and finetuned with LDAM+CP loss. Here, in the main text, we summarize our findings. See Subsection \ref{ss:qualitative analysis details} for more details.

In several cases, the model's predictions indicate  manual annotation mistakes. Moreover, some potential sources for model misclassification are 
\begin{inparaenum}
    \item data scarcity
    \item inadequate context window
    \item some very interesting patterns that -- according to our hypothesis -- do not match previous patterns from the source domain regarding how translators are represented.
\end{inparaenum}
Again, see Subsection A.2 for more details.

\section{Conclusion}

We study 2 complex coding systems which were developed on paragraphs of a Hungarian literary journal to track trends in the social perception of literary translation: a multilabel content label set, and a multiclass context label set. Although both label sets are highly imbalanced, we show that with extensive hyperparameter tuning and loss functions robust to imbalance it is possible to teach BERT models both label sets. This result is verified with 10-fold crossvalidation. We further investigate if the resulting ensemble of models is capable of carrying over the coding systems to another literary journal. To that end, we introduce a method to estimate the confidence interval of evaluation results on a test set sampled on the target domain with a given sample size. With this, we verify that our ensemble of models can fulfill this task in the case of content labels. We conduct numerous comparisons to low-cost baseline methods and variations in our BERT training procedure. In particular, we show that domain adaptation to OCR-ed text of distinct subject matter already significantly helps task performance.

\section{Acknowledgements} The authors would like to thank Gábor Berend, Adrián Csiszárik, Péter K\H orösi-Szabó, Anikó Sohár and Dániel Varga for their insight and many helpful discussions. We are also grateful to the reviewers for their valuable remarks and suggestions. Zsámboki is supported by the Ministry of Innovation and Technology NRDI Office within the framework of the Artificial Intelligence National Laboratory (RRF-2.3.1-21-2022-0004) and the ELTE TKP 2021-NKTA-62 funding scheme.

\section{Ethical Statement: Carbon Footprint}

We estimate that in total, experiments related to this project took 16 months, that is 11520 NVIDIA A100 40GB GPU hours. The Machine Learning Emissions Calculator by \citet{lacoste2019quantifying} estimates that this emitted 1244.16 kg CO${}_{2}$eq. Based on data published in Our World in Data \citep{carbontravel}, travelling as a passenger 7318.59 km in a car, 6220.8 km on a flight or 12826.39 km by bus would emit a similar amount.


\nocite{*}
\section{Bibliographical References}\label{sec:reference}

\bibliographystyle{lrec-coling2024-natbib}
\bibliography{lrec-coling2024-example}

\section{Language Resource References}
\label{lr:ref}
\bibliographystylelanguageresource{lrec-coling2024-natbib}
\bibliographylanguageresource{languageresource}

\appendix

\section{Label Examples}\label{s:label examples}

Below, we give the definition of two of the labels, the content label \emph{author as translator} and the context label \emph{translator}. Afterwards, we provide the details of the qualitative analysis that was described in Subsection \ref{ss:qualitative analysis}.

\subsection{Definitions}\label{ss:label definitions}

The content label \emph{author as translator} means that in the paragraph, somewhere a translator is mentioned who is more known by their work as an author. There can be many other content labels applied to the paragraph in question, the part labeled \emph{author as translator} can be a minor detail and it is also important that it is about the fact that the person mentioned is more famous about something other than translation.

The context label \emph{translator} on the other hand means that the \emph{reason} translation in that paragraph is mentioned is that the  topic is a specific translator for any reason at all. The depth of the discussion, other themes or the way translation is mentioned are irrelevant here.

There can be correlation between these two labels but in spite of the similar themes they explore, they do not necessarily occur together and there is no overlap in their function.

\subsection{Qualitative Analysis Details}\label{ss:qualitative analysis details}

\subsubsection{Author as Translator}

For this binary label, we view a paragraph as positive if the average label probability by the model ensemble is larger than 50\%. This holds for 18 paragraphs out of 100. Out of these 18 misclassifications, 6 turn out to be a mistake in manual annotation. This gives a strong indication as to how helpful our tool can be for annotation.

Of the rest, in 2 false negative cases we hypothesize that it is mostly the name of the translator  that indicates the validity of the label and it is a relatively less well-known name that might not have been frequently present in the rest of the corpus.

We also noticed that in 4 false positive cases, the translator in question is most known for their work as a translator, however, the discourse exhibits traits rarely displayed without the \emph{author as translator} label in the corpus according to our hypothesis based on the train set. These are the following: 
\begin{inparaenum}
    \item details of the translator being famous, i.e.~winning prizes and having a prestigious portfolio and
    \item writing extensively/being interviewed about themselves or their practice.
\end{inparaenum}

What is even more intriguing is that each of these four instances are about translators who work from Hungarian. We very tentatively pose the hypothesis that translating from Hungarian is an activity that could be viewed more important in the Hungarian discourse because it leads to the visibility and representation of Hungarian literature and that could be a more personal matter in the Hungarian literary field than the accessibility and representation of foreign literature in Hungary. This, however, requires further investigation.

\subsubsection{Translator}

Here, out of the 15 misclassifications, 2 turned out to be manual annotation mistakes. Of the rest, in 3 cases, it is possible that the broader context cannot be inferred from the paragraph and would require knowledge of a larger or different portion of the text. In another 3 cases, although the paragraph does extensively feature the translation activity of a single individual, the broader context is not about the translator.

\section{Hyperparameter search spaces}

In the following table, the hyperparameter initial distributions used in the domain adaptation BlendSearch (\S \ref{subsection: domain adaptation}) and finetuning Population-Based Training (\S \ref{subsubsection:population-based training}) are provided. For an interval $I\subseteq\mathbf R$, let $\ell\mathcal U$ denote the log uniform distribution $\exp(\mathcal U\log(I))$, and let $d\ell\mathcal UI$ denote the discrete log uniform distribution $\lfloor\ell\mathcal UX\rfloor$.

\begin{table*}[!ht]
\centering
\begin{tabular}{cc}
    \toprule
    Hyperparameter & Initial Distribution \\
    \midrule
    AdamW first moment, $\beta_1$ in  \citep[Algorithm 2]{loshchilov2019decoupled}$^{123}$ & $1-\ell\mathcal U[10^{-2},\frac12]$ \\
    AdamW second moment, $\beta_2$ in  \citep[Algorithm 2]{loshchilov2019decoupled}$^{123}$ & $1-\ell\mathcal U[10^{-4}, 10^{-1}]$ \\
    Complexity, $C$ in \citep[Equation 11]{cao2019learning}$^3$ & $\ell\mathcal U[10^{-2}, 10^2]$ \\
    Confidence penalty strength, $\beta$ in \citep[\S3]{pereyra2017regularizing}$^3$  & $\ell\mathcal U[10^{-2}, 10^2]$ \\
    Focusing parameter, $\gamma$ in \citep[Equation 4]{lin2017focal}$^2$ & $\ell\mathcal U[2^{-4}, 2^4]$ \\
    Learning rate$^{123}$ & $\ell\mathcal U[10^{-5}, 10^{-2}]$ \\
    Learning Rate scheduler (after warmup)$^1$ & $\mathcal U\{\mathrm{constant}, \mathrm{cosine}, \mathrm{linear}\}$ \\
    Maximum gradient norm$^{123}$ & $\ell\mathcal U[10^{-2}, 10^{2}]$ \\
    Maximum train epochs$^1$ & $d\ell\mathcal U[1, 101)$ \\
    Weight decay rate, $\lambda$ in \citep[Algorithm 2]{loshchilov2019decoupled}$^{123}$ & $\ell\mathcal U[10^{-1}, 1]$ \\
    
    \bottomrule

\end{tabular}
\caption{Hyperparameter initial distributions. $^1$Used in domain adaptation. $^2$Used in content label finetuning. $^3$Used in context label finetuning.}
\label{table:hyperparameter initial distributions}

\end{table*}

\section{Constituents of the OCR Corpus}\label{s:ocr corpus}
The OCR corpus consists of the 1980--1999 issues of the following journals:
\begin{enumerate}
    \item \emph{Akadémiai Közlöny} (later \emph{Akadémiai Értesítő}) was the bulletin of the Hungarian Academy of Sciences during the period under examination.
    \item \emph{Állam és Igazgatás} (later \emph{Magyar Közigazgatás}) was a social studies journal specializing in public administration.
    \item \emph{Gyermekgyógyászat} was the journal of the Pediatrists Group in the Medical and Sanitary Workers Union.
\end{enumerate}

\end{document}